\documentclass{ecai2006}
\usepackage{times,algorithm2e,epsfig,amssymb,bm}
\usepackage{graphicx}
\usepackage{latexsym}
\usepackage{amsmath}
\usepackage{amsfonts,eufrak,yfonts}
\usepackage{url}

\newtheorem{lemma}{Lemma}

\newcommand{\pr}{\mbox{\textbf{Pr}}}
\newcommand{\tr}{\mbox{\textsl{tr}}}
\newcommand{\mc}{$\mathfrak{M}$}

\def\cqfd{\hfill\hbox{$\hbox{\vrule width 0.8pt
\vbox to6pt{\hrule depth 0.8pt width 5.2pt
\vfill\hrule depth 0.8pt}\vrule width 0.8pt}$}}

\begin{document}

\title{Soft Uncoupling of Markov chains for Permeable Language Distinction: A New Algorithm}

\author{Richard Nock$^{1}$ Pascal Vaillant$^{1}$, Frank Nielsen$^{2}$, Claudia Henry$^{1}$\\
  \mbox{~}\\
  (1) GRIMAAG, Universit{\'{e}} des Antilles-Guyane, 97233 Sch{\oe{}}lcher, Martinique\\
  (2) Sony CSL, 3-14-13 Higashi Gotanda, Shinagawa-Ku. 141-0022 Tokyo, Japan\\
  \mbox{~}\\
  \texttt{\{rnock,pvaillan\}@martinique.univ-ag.fr}
}

\date{May 9$^{th}$, 2006}

\maketitle
 
%\pagenumbering{arabic}
%\setcounter{page}{1}%Leave this line commented out.

\begin{abstract} \small\baselineskip=9pt
Without prior knowledge, distinguishing different languages 
may be a hard task, especially when their borders are permeable.

We develop an extension of spectral clustering --- a powerful unsupervised classification 
toolbox --- that is shown to resolve accurately the task of soft
language distinction. At the heart of our approach, we replace the
usual hard membership assignment of spectral clustering by a soft, probabilistic
assignment, which also presents the advantage to bypass a well-known complexity bottleneck
of the method. 
Furthermore, our approach relies on a novel, convenient construction of a 
Markov chain out of a corpus. Extensive experiments with a readily available system
clearly display the potential of the method, which brings a visually appealing soft 
distinction of languages that may define altogether a whole corpus.
\end{abstract}

\section{Introduction}

This paper is concerned with unsupervised learning, the task that
consists in assigning a set of objects to a set of $q>1$ so-called
clusters. For the purpose of text classification, we suppose that
objects are words: each cluster should define a set of words which are
syntactically close to one another, while different clusters should be
as different as possible from the syntactic standpoint.

There are two main ways of understanding what is meant by
``syntactically close'': following~\cite{sCLG}, it is generally
acknowledged that two combined ``axes'' define the combinatorial
possibilities of a language's syntax: the {\em syntagmatic axis} is
the linear dimension of the text, where occurrences of words actually
appear one after another; the {\em paradigmatic axis} is the dimension
of all possible alternative choices available at a given position to a
speaker or writer. Hence, two words are ``syntactically close'' to one
another on the syntagmatic axis, if they often tend to appear
together, at specific relative positions, in common contexts; they are
close to one another on the paradigmatic axis, if they appear
alternatively in similar positions within comparable contexts. The
first criterion defines a measure of word distance within a text, and
is suited to studying problems such as internal coherence of text
segments \cite{uiTS}; the second defines a measure of word distance
within a class, and is suited to studying problems like defining
relevant syntactic \cite{sPOS,bgUV} or semantic
\cite{ptlDC,brcTS,dlpSM} categories. In the frame of this paper,
attention will be drawn on the first one of these problems, which has
not been very extensively tackled.

A challenging application in the field of linguistic engineering is
language identification and comparison. Language identification for
itself is now considered an easy task on monolingual text documents,
as two very reliable methods (based on frequency analyses on the most
frequent words, and on the most frequent $n$-byte sequences) may be
mixed to get optimal results \cite{gLIS}; yet some work remains to be
done for the task of language identification on multilingual
documents, where a non-trivial question is the definition on language
section boundaries \cite{vIST}. This question is particularly
interesting when the border between different languages is
permeable. This is typically the case within the group of Creole
languages of the Caribbean region.

Creole languages in their present form have emerged during a short
period of time (probably less than one century, in the late 17th
century), in very atypical conditions of language transmission and
evolution. They have developed in the newly colonized West Atlantic
territories (in the Caribbean islands and on both American mainlands),
on the basis of Western European languages spread by the nations most
involved in colonization (French, English, Portuguese and Dutch), but
in sociolinguistic situations where, due to the rapidly growing slave
trade economy, there could be, within every single generation, less
than 50\% of native speakers of the language in its current state of
development involved in the speaking community. Even in periods of
fast language evolution (like for the case of Middle English between
the 11th and 15th century), no European language has experienced such
a phase of ``linguistic stress''. After the 18th century, the language
situation has somehow stabilized, although Creoles still undergo
linguistic change at a pace which is probably faster than many well
established languages.

In at least some cases, the Creole language has remained in contact
with its ``lexifier'' European language (none of those has in the
meantime become extinct), in sociolinguistic situations which have
sometimes been coined as ``diglossic'': this has especially been the
case for English-based Creoles like Jamaican or Gullah (spoken in the
USA states of South Carolina and Georgia); and, closer to our study's
main focus, for French-derived Creoles spoken on the territories of
Haiti, Guadeloupe, Martinique and French Guiana. In a diglossic
situation, the European language is still in use as the official and
prestige language, while the Creole language is the vernacular. This
leads to very frequent code-switching and to intermingling of
languages in several domains. Thus, when it comes to corpora of
linguistic productions coming from this type of speech community, the
question of the ``border'' between languages can be asked on two
distinct planes: on the plane of structural (merely linguistic)
properties, and on the plane of the situations of use.

The first question involves problems of language clustering. A
learning task might consist in drawing a cladogram (family tree) of
various French-based Creole languages on the basis of their structural
similarities. Studying ``paradigmatic'' syntactic closeness (i.e.
context similarity, see above), might also help define the most
appropriate part-of-speech categorization for those languages, and
check the appropriateness of eurocentric grammatical descriptions in
their case. But this is not the main scope of the present paper.

The second question involves delimiting the use of every language
in multilingual texts or speech productions, and this is the task
on which we will now concentrate.

% Fin paragraphe Pascal

In the last few years, the most prominent developments of text classification
have concerned \textit{supervised} classification (\textit{i.e.} texts have
explicit labels to predict), with the advent of algorithms powerful enough to 
process texts described with the simplest conventions (\textit{e.g.} attribute-value vectors) 
\cite{jTC,jTI,ssBT}. A glimpse at its unsupervised side easily reveals that 
classification has so far comparatively remained quite distant from text classification,
at least for its most recent breakthroughs in learning / mining. \textit{Spectral clustering}
is a very good example, with such a success that its recent developments
have been qualified elsewhere as a ``gold rush'' in classification 
\cite{dAT,bjLS,bgUV,bnLE,msLS,mSA,njwOS} (and many others), pioneered by works in spectral
graph theory \cite{cSG} and image segmentation \cite{smNC}.
Roughly speaking, spectral clustering consists in finding some principal axes of a similarity
matrix. The subspace they span, onto which the data are projected, \textit{may} yield clusters 
optimizing a criterion that takes into account both the maximization of the within-cluster 
similarity, and the minimization of the between-clusters similarity. Among the
attempts to cast spectral clustering to text classification, one of the first
builds the similarity matrix via the computation of cosines between
vector-based representations of words, and then builds a normalized 
graph Laplacian out of this matrix to find out the principal axes \cite{bgUV}. 

The papers that have so far investigated spectral clustering have two commonpoints.
First, they consider a \textit{hard} membership assignment of data: the
clusters induce a partition of the set of objects. It is widely known that \textit{soft}
membership, that assigns a fraction of each object to each cluster, is sometimes
preferable to improve the solution, or for the problem at hand. This
is clearly our case, as words may belong to more than one language cluster.
In fact, this is also the case for the probabilistic (density estimation) approaches to clustering,
pioneered by the popular Expectation Maximization \cite{dlrML}.
Their second commonpoint is linked to the first: the solution of clustering
is obtained \textit{after} thresholding the spectral clustering output. This
is crucial because in most (if not all) cases, the optimization of the clustering 
quality criterion is \textit{NP-Hard} for the hard membership assignment \cite{smNC}.
To be more precise, the principal axes yield the polynomial time \textit{optimal} solution to
an optimization problem whose criterion is the \textit{same} as that of hard membership 
(modulo a constant factor),
\textit{but} whose domain is unconstrained. Hard membership makes it necessary to 
fit (threshold) this optimal solution to a constrained domain. Little is currently 
known for the quality of this approximation, except for the NP-Hardness of the task.

This paper, which also focuses on spectral clustering, departs from the mainstream for
the following reasons and contributions. First (Section 2), compared to text classification
approaches, we do not build the similarity matrix in
an \textit{ad hoc} manner like \cite{bgUV}. Rather, we consider that the corpus is
generated by a stochastic process following a popular bigram model \cite{psCN}, out of
which we build its maximum likelihood Markov chain. This particular Markov chain satisfies all 
conditions for a convenient spectral decomposition. Second (Section 3), we propose an extension of
spectral clustering to \textit{soft spectral clustering}, for which we give
a probabilistic interpretation of the spectral clustering output. Apart from our
task at hand, which justifies this extension, we feel that such results may be of
independent interest, because they tackle the interpretation of the \textit{tractable}
part of spectral clustering, avoiding the complexity gap that follows after
hard membership. Last (Section 4), we provide experimental results of soft spectral
clustering on a readily available system; experiments clearly display the potential of this 
method for text classification.

\section{Maximum likelihood Markov chains}

In this paper, calligraphic faces such as ${\mathcal{X}}$ denote sets and blackboard 
faces such as ${\mathbb{S}}$ denote subsets of ${\mathbb{R}}$, the set of real numbers; 
whenever applicable, indexed lower cases such as $x_i$ ($i=1, 2, ...$) enumerate the 
elements of ${\mathcal{X}}$. Upper cases like $M$ denote matrices, with $m_{i,j}$ being 
the entry in row $i$, column $j$ of $M$; $M^{\top}$ is the transposed of $M$. Boldfaces such as $\bm{x}$ denote column vectors, with $x_i$ being the $i^{th}$ 
element of $\bm{x}$. 
A corpus ${\mathcal{C}}$ is a set of texts, 
$\{{\mathcal{T}}_1, {\mathcal{T}}_2, ..., {\mathcal{T}}_m\}$, with $m$ the length of the corpus. 
$\forall 1\leq k\leq m$, text 
${\mathcal{T}}_k$ is a string of tokens (words or punctuation marks), 
${\mathcal{T}}_k=\omega_{k,1} \omega_{k,2} ... 
\omega_{k,|{\mathcal{T}}_k|}$, of size $|{\mathcal{T}}_k|$, with $|.|$ the cardinal (whole
number of tokens of ${\mathcal{T}}_k$). The \textit{size} of
the corpus, $|{\mathcal{C}}| = n$, is the sum of the
length of the texts: $n = \sum_{i=1}^m {|{\mathcal{T}}_i|}$. 
The size of a corpus is implicitly
measured in words, but it may contain punctuation marks as well.
The \textit{vocabulary} of ${\mathcal{C}}$, ${\mathcal{V}}$, is the
set of distinct linguistically relevant words or punctuation marks,
the tokens of which are contained in the texts of ${\mathcal{C}}$.
The size of the vocabulary is denoted $v=|{\mathcal{V}}|$. The elements of ${\mathcal{V}} = \{v_1, v_2, ..., v_v\}$ are \textit{types}:
each one is unique and appears only once in ${\mathcal{V}}$. $\forall i,j \in \{1, 2, ..., v\}$, we let $n_i$ denote the number of occurrences of type $i$ in ${\mathcal{C}}$, and $n_{i,j}$ the number of times a word of type $i$ immediately preceeds (left) a word of type $j$ in ${\mathcal{C}}$.
Finally, we denote \mc~a (first order) Markov chain, with state space ${\mathcal{V}}$, and transition probability matrix $P_{v \times v}$. $P$ is row stochastic: $p_{i,j} \geq 0$ ($1\leq i,j\leq v$) and $\sum_{j=1}^{v} {p_{i,j}=1}$ ($1\leq i\leq v$).
Suppose that ${\mathcal{C}}$ is generated from \mc. The most natural way to build $P$ is to maximize its likelihood with respect to ${\mathcal{C}}$. The solution is given by the following folklore Lemma.
\begin{lemma}\label{theo1}
The maximum likelihood transition matrix $P$ is defined by $p_{i,j} = n_{i,j} / n_i$, with $1\leq i,j\leq v$.
\end{lemma}
The computation of $P$ as in Lemma \ref{theo1} is convenient
\textit{if} we make the assumption that a text is written from the
left to the right. This corresponds to an {\em a priori} intuition of
speakers of European languages, who have been taught to read and write
in languages where the graphical transcription of the linearity of
speech is done from left to right. However, a more thorough reflection
on the empirical nature of the problem has lead us to question this
approach.  The method being developed should be able to work on any
type of written language, making no assumption on its transcription
conventions. Some languages (among which important literary languages
like Hebrew or Arabic) have a tradition of writing from right to left,
and this sometimes goes down to having the actual stream of bytes in
the file also going ``from right to left'' (in the file access sense).
The new Unicode standard for specifying language directionality
circumvents this, by allowing the file to always be coded in the
logical order, and managing the visual rendering so that it suits the
language conventions, even in the case of mixed-language texts (i.e.
English texts with Hebrew quotes); but large corpora still are encoded
in the old way, and the program should not be sensitive to this. More
generally, the method we propose should be designed to accept any file
as a statical, empirical object, and should be able to find laws and
regularities in it, making no more postulates than necessary.

\begin{figure}
\centerline{\input{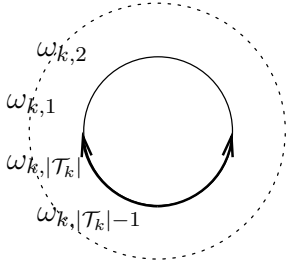}}
\caption{A ``circular'' generation of a text ${\mathcal{T}}_k$ eliminate both the direction for writing ${\mathcal{C}}$ (arrows) \textit{and} the choice of the first word written.\label{f-circ}}
\end{figure}

We have found a convenient approach to eliminate this directionality
dependence. It also has the benefit of removing the dependence in the
choice of the first word to write down a text. Everything is like if
we were computing the likelihood of ${\mathcal{C}}$ with respect to
the writing of its texts in a \textit{circular} way. Figure
\ref{f-circ} presents the writing of text ${\mathcal{T}}_k$: we pick a
random word, and then move either clockwise or counter clockwise
to write words. After we have made a complete turn, everything is like
if we had written twice ${\mathcal{T}}_k$.
The following Lemma, whose proof is direct from Lemma \ref{theo1}, gives the new maximum likelihood transition matrix $P$ (proof omitted to save space).
\begin{lemma}\label{theo2}
With the circular writing approach, $P = D^{-1} W$, with $W_{v \times v}$ such that $w_{i,j} = (n_{i,j} + n_{j,i})/2$, and $D_{v\times v}$ diagonal with $d_{i,i} = d_i = n_i$.
\end{lemma}
From now on, we use the expression for $P$ in Lemma \ref{theo2}. The
circular way to write down the texts of ${\mathcal{C}}$ has another
advantage: \mc~is irreducible. Let us make the assumption that \mc~is
also aperiodic. This derives from a clearly mild assumption, namely
that \mc~satisfies for a vocabulary large enough, as in this case
loops of arbitrary long size tend to appear between
words. Irreducibility and aperiodicity imply that \mc~is ergodic,
\textit{i.e.} regardless of the initial distribution, \mc~will settle
down over time to a single stationary distribution $\bm{\pi}$ solution
of $P^{\top} \bm{\pi} = \bm{\pi}$, with $\pi_i = n_i / n$ \cite{msLS}.

\section{From hard to soft spectral clustering}

Fix $q > 1$ some user-fixed integer that represents the number of clusters to find.
The ideal objective would be to find a mapping $Z: {\mathcal{V}} \rightarrow {\mathbb{S}}^q$,
with ${\mathbb{S}} = \{0,1\}$, mapping that we represent by a
matrix $Z = [\bm{z}_1, \bm{z}_2, ..., \bm{z}_q] \in {\mathbb{S}}^{v \times q}$.
Under appropriate constraints, the mapping should minimize a \textit{multiway normalized cuts}
(MNC) criterion \cite{bjLS,bgUV,bnLE,msLS,mSA,njwOS,smNC}:
\begin{eqnarray}
\arg\min_{Z \in {\mathbb{S}}^{v \times q}} \mu(Z) & = & \sum_{k=1}^{q} {\kappa_k(Z) / \alpha_k(Z)} \:\:, \label{defmu}\\
& \mbox{s.t.} & Z^\top Z \mbox{ positive diagonal} \nonumber \\
& \mbox{s.t.} & \tr(Z^\top Z) = v \nonumber \:\:,
\end{eqnarray}
with $\kappa_k(Z) = \sum_{i,j=1}^{v} {w_{i,j} (z_{i,k} - z_{j,k})^2}$ and $\alpha_k(Z) = \sum_{i=1}^{v} {z^2_{i,k} d_i}$. Since this does not change the value of $\mu(Z)$, we suppose without loss of generality that $w_{i,i} = 0, \forall 1\leq i\leq v$. Because of the constraints on $Z$ in (\ref{defmu}), it induces a natural hard membership assignment on $\mathcal{V}$ (\textit{i.e.} a partition), as follows:
\begin{eqnarray}
{\mathcal{V}}_k & = & \{v_i : z_{i,k} = 1\}\:\:, \forall 1\leq k\leq q \:\:. \label{eqv}
\end{eqnarray} 
There is one appealing reason why clustering gets better as MNC in (\ref{defmu}) is minimized. Suppose we start (at $t=0$) a random walk with the Markov chain \mc, having transition matrix $P$, and from its stationary distribution $\bm{\pi}$. Let $[{\mathcal{V}}_k]_t$ be the event that the Markov chain is in cluster $k$ at time $t\geq 1$. We obtain the following result \cite{msLS}: 
\begin{eqnarray}
\mu(Z) & = & 2 \sum_{k=1}^{q} {\pr([\overline{{\mathcal{V}}_k}]_{t+1} | [{\mathcal{V}}_k]_t)} \label{theoms}
\end{eqnarray}
for the partition defined in eq. (\ref{eqv}). Thus, $\mu(Z)$ sums the probabilities of escaping a cluster given that the random walk is located inside the cluster: minimizing $\mu(Z)$ amounts to partitioning ${\mathcal{V}}$ into ``stable'' components with respect to \mc.
Unfortunately, the minimization of MNC is NP-Hard, already when $q=2$ \cite{smNC}. To approximate this problem, the output is relaxed and the goal rewritten as seek:
\begin{eqnarray}
\arg\min_{Y \in {\mathbb{R}}^{v \times q}} \nu(Y) & = & \sum_{k=1}^{q} {\kappa_k(Y)} \:\:, \label{defnu}\\
& \mbox{s.t.} & Y^\top D Y = I \nonumber \:\:.
\end{eqnarray}
This problem is tractable by a spectral decomposition of \mc~(see \textit{e.g.} \cite{smNC}), which yields that $Y$ is the set of the $q$ column eigenvectors associated to the smallest eigenvalues of the generalized eigenproblem ($\forall 1\leq k\leq q$):
\begin{eqnarray}
(D-W) \bm{y}_k & = & \lambda_k D \bm{y}_k \:\:, \label{eqeig}
\end{eqnarray}
and it comes $\nu(Y) = 2 \sum_{k=1}^{q} {\lambda_k}$. If we suppose, without loss of generality, that eigenvalues are ordered, $\lambda_1 \leq \lambda_2 \leq ... \leq \lambda_q$, then it easily comes $\lambda_1 = 0$, associated to a constant eigenvector $\bm{y}_1$ \cite{smNC}. People usually discard this first eigenvector, and keep the following ones to compute $Z$ after a heuristic thresholding of $Y$. The proof that this thresholding is heuristic follows from the fact that if we restrict (\ref{defnu}) to thresholded matrices (whose rows come from a set of at most $q$ distinct row vectors), then it becomes equivalent to (\ref{defmu}), \textit{i.e.} intractable \cite{bjLS}.\\

Notice however that the spectral relaxation finds the \textit{optimal} solution to (\ref{defnu}) in time $O(q v^3)$ (without algorithmic sophistication), from which the heuristic thresholding only aims at recovering a hard membership assignment. Whenever a soft membership assignment is preferable, we show that one can be obtained directly from $Y$, which is optimal with respect to a criterion similar to (\ref{theoms}), while its computation bypasses the complexity bottleneck of hard membership, thus killing two birds in one shot.\\
\begin{figure*}[ht]
\begin{center}
\begin{tabular}{|c|c|}\hline
\epsfig{file=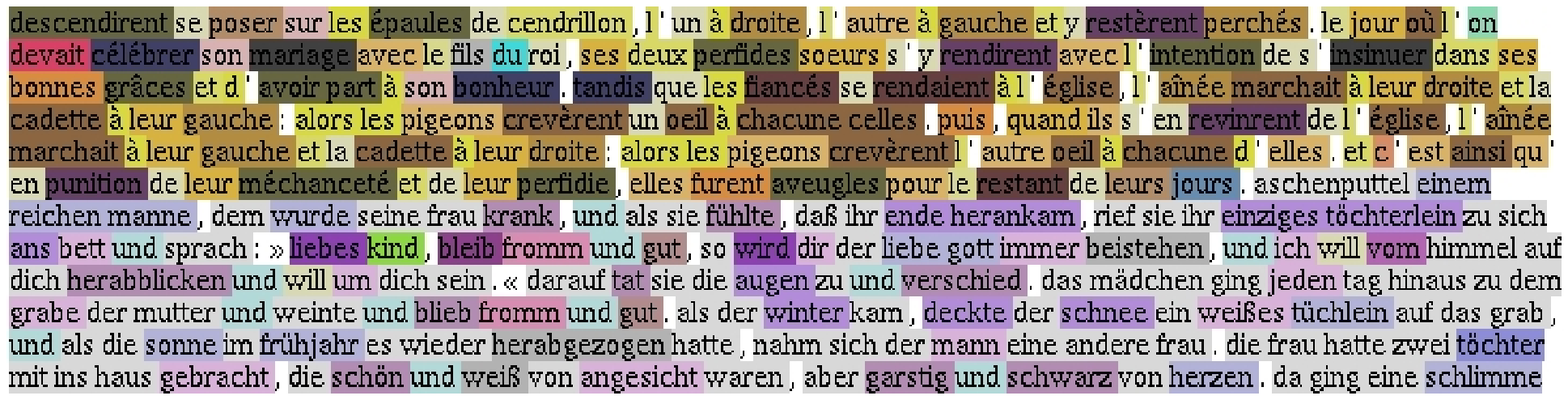,width=7.50cm} & \epsfig{file=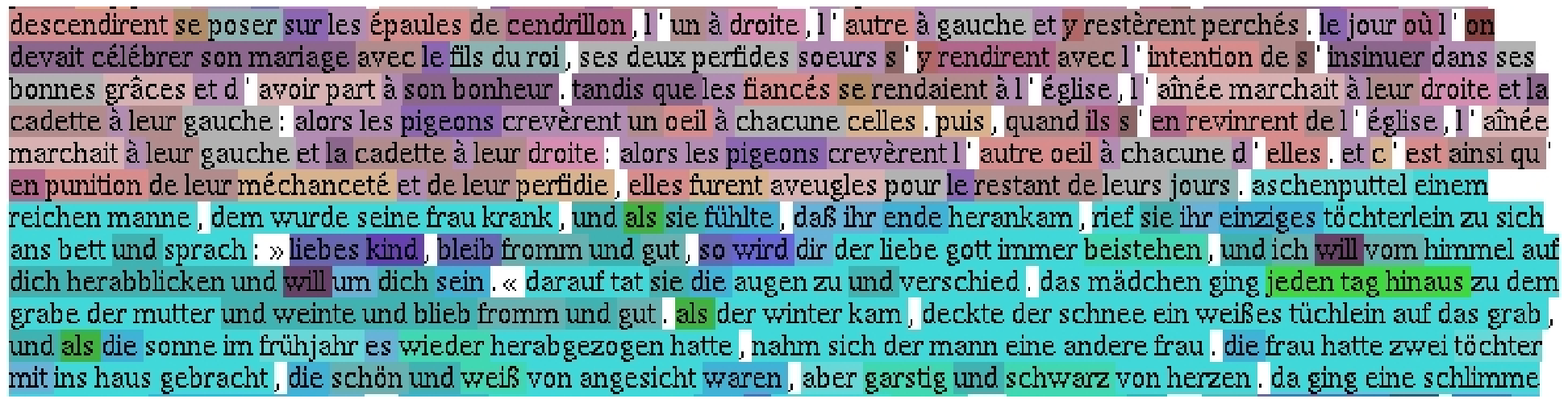,width=7.50cm} \\
\epsfig{file=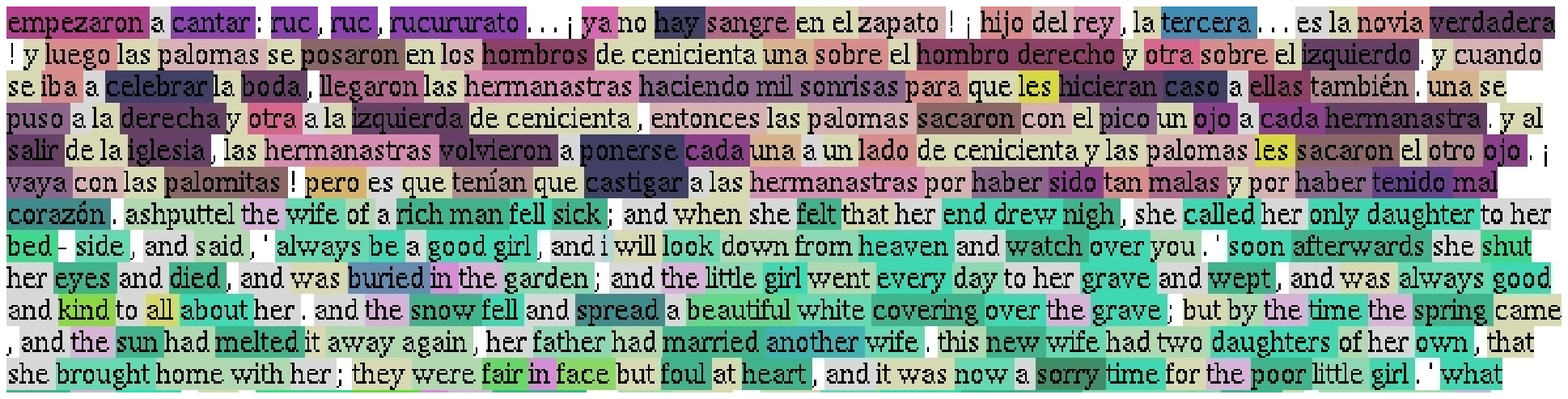,width=7.50cm} & \epsfig{file=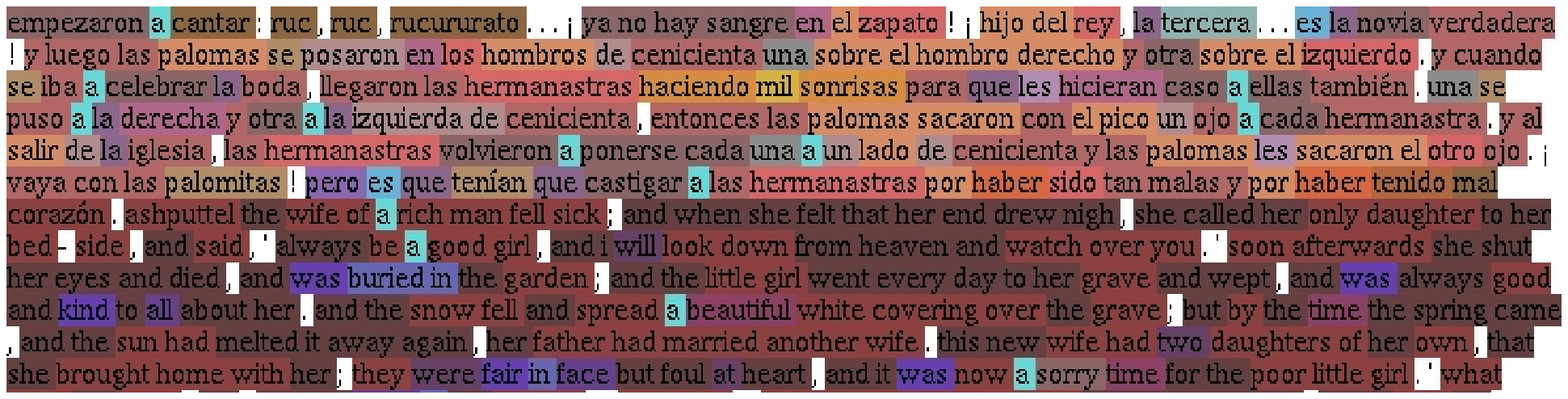,width=7.50cm}\\ \hline
$\pr([v_i]_t | [{\mathcal{V}}_k]_t) = d_i y^2_{i,k}$ & $y_{i,k}$\\ \hline
\end{tabular}
\end{center}
\caption{Experiments on multilingual passages of {\em Cinderella}. Each row crops a borderline between two languages (from the top to the bottom): French / German, Spanish / English. The bottom row displays the quantities that are represented by RGB colors in each column, where each color level is associated to a principal axis $k \in 2, 3, 4$.\label{f-cind}}
\end{figure*}
For this purpose, define matrix $\tilde{Y}$ from $Y$ as:
\begin{eqnarray}
\tilde{y}_{i,k} & = & d_i y^2_{i,k} \label{defcond} \:\:.
\end{eqnarray}
Then, we have $\tilde{Y}^\top \bm{1} = \bm{1}$,
\textit{i.e.} each column vector $\tilde{\bm{y}}_k$ of $\tilde{Y}$ defines
a probability distribution over ${\mathcal{V}}$.
Since $\tilde{\bm{y}}_k$ is associated to principal axis $k$, it seems natural to define it as the probability to draw $v_i$ given that we are in ${\mathcal{V}}_k$, the cluster associated to the axis. Following the notations of eq. (\ref{theoms}), we thus let:
\begin{eqnarray}
\tilde{y}_{i,k} & = & \pr([v_i]_t | [{\mathcal{V}}_k]_t) \label{defprob}
\end{eqnarray}
be the probability to pick type $v_i$, given that we are in cluster $k$, at time $t$. This is our
 soft membership assignment: axes define clusters, and the column vectors of $\tilde{Y}$ define
the distributions associated to each cluster. Notice that this provides us with a sound extension
of the hard MNC solution (\ref{eqv}) for which $\tilde{y}_{i,k}$ equals 1 for a single cluster,
and zero for the other clusters ($\forall 1\leq i\leq v$). We also have more, as this brings a direct
and non trivial generalization of (\ref{theoms}).
Define matrix $P^{(k)}$ such that $p^{(k)}_{i,j} = (w_{i,j}y_{j,k})/(d_i y_{i,k})$. $p^{(k)}_{i,j}$ is akin to the difference between the probabilities of reaching respectively ${\mathcal{V}}_k$ and $\overline{{\mathcal{V}}_k}$ in $j$, \textit{given} that the random walk is located on $i$ ($\forall t\geq 0$): $p^{(k)}_{i,j} = \pr([v_j \wedge {\mathcal{V}}_k]_{t+1} | [v_i]_t ) - \pr([v_j \wedge \overline{{\mathcal{V}}_k}]_{t+1} | [v_i]_t )$.
Provided we make the assumption that reaching a type outside cluster $k$ at time $t+1$ does not depend on the starting point at time $t$, an assumption similar to the memoryless property of Markov chains, we obtain our main result, whose proof relies on applications of Bayes rules.
\begin{theorem} $\nu(Y) = 4 \sum_{k=1}^{q} {\pr([\overline{{\mathcal{V}}_k}]_{t+1} | [{\mathcal{V}}_k]_t)}$.
\end{theorem}
By means of words, solving (\ref{defnu}) brings the soft clustering whose components have \textit{optimal} stability, and whose associated distributions are given by $\tilde{Y}$. As a consequence, we easily obtain that $\tilde{\bm{y}}_1 = \bm{\pi}$, 
the stationary distribution. This is natural, as this is 
the observed distribution of types, \textit{i.e.} the one
 that best explains the data. In previous results, 
\cite{bgUV} choose the Brown corpus and
make a 2D plot of some spectral clustering results on the second and
third principal axes, \textit{after} having made a prior selection of
the most frequent words (to be plotted). From $\tilde{\bm{y}}_1$, it
comes that this amounts to make a selection of words according to the
\textit{first} principal axis, which is not plotted. 

\section{Experiments}

A computer program has been developed to implement word classification
and text segmenting according to the method explained above. It is
publicly available through a {\sc cgi}\footnote{{\sc url}:
\url{http://www.univ-ag.fr/~pvaillan/mots/}}. The program takes a
text of arbitrary long size as input. First, it automatically detects
the text format and encoding, and converts everything to raw text
encoded in Unicode UTF-8. Second, it performs a stage of tokenization,
i.e. it segments the raw stream of bytes into tokens of words, figures
or typographical signs. Third, it builds an index table suited for
fast access to word type information (designed on the lexical tree, or
{\em trie}, model). Fourth, it computes the bigram transition matrix
$T$ ($n_{i,j}$) (lemma~\ref{theo1}), by moving a contextual window
along the tokens put in their text order, and incrementing $n_{i,j}$
for every seen occurrence of a transition ($\omega_{i}$,$\omega_{j}$);
$W$, and then $P$ (as given by lemma~\ref{theo2}) are then computed
from $T$. Fifth, it makes use of the linear algebra functions of the
{\sc lapack} library\footnote{{\sc lapack url}:
\texttt{http://www.netlib.org/lapack/}.} to compute the eigenvalues
and eigenvectors of the matrices.

The program's results are displayed in a way designed to give the user
a visual representation of every word's soft membership to the
clusters. For this purpose, we can represent each word with a RGB
color, where each color level is associated to some principal axis
$k$, and scales the component of $\tilde{y}_{i,k}$ for each word
$i$. This allows the choice of three axes to compute the color. Let us
assume we want to be able to display $\chi{}$ different color levels
on each axis (in our illustrations, $\chi{}=5$); For every selected
component $k$, the $v$ different values for $\tilde{y}_{i,k}$ are
grouped into $\chi{}$ connected intervals $I_{1}$, $I_{2}$~\dots
$I_{\chi{}}$, not necessarily of the same length, such that
$\cup_{i=1}^{\chi{}} I_{i} = [0,1]$, and such that every interval
contains the same number of points (approximately $v/5$): this yields
the maximum visual contrast.

Figure \ref{f-cind} presents such an experiment on a 1Mb text,
containing four versions of the same tale (Cinderella, from the Grimm
Brothers), in four languages: French, German, Spanish, and English. We
have plotted both $\tilde{\bm{y}}_k$ (left column) and $\bm{y}_k$
(right column) for each word, removing punctuation marks from the
spectral analysis (this explains why they are displayed in white).

Both columns show that the representations manage to cluster all
languages. The right column also gives access to the sign of
$\bm{y}_{k}$ (in this case $\cup_{i=1}^{\chi{}} I_{i} = [-1,1]$), while
values for the left colum, $\pr([v_i]_t | [{\mathcal{V}}_k]_t) = d_i y^2_{i,k}$,
belongs to a smaller interval, $[0,1]$. In this context, it 
is quite interesting to see that the contrasts between languages is 
marked on both columns. What is most interesting in this context is that 
the contrast inside each language is actually sharper for 
$\pr([v_i]_t | [{\mathcal{V}}_k]_t)$. While the colors
distinguish the languages, they also ``order'' them in some sense.
From the average color levels of each language,
we can say that R(ed) is principally German, G(reen) is principally 
English, and
B(lue) is principally Spanish. French is somewhere in between all of them. 
What is much interesting
is that all this is in good accordance with the roots of these four languages,
a fact which is of course utterly out of sight for the computer program.

\begin{figure*}[ht]
\begin{center}
\begin{tabular}{|c|c|}\hline
\epsfig{file=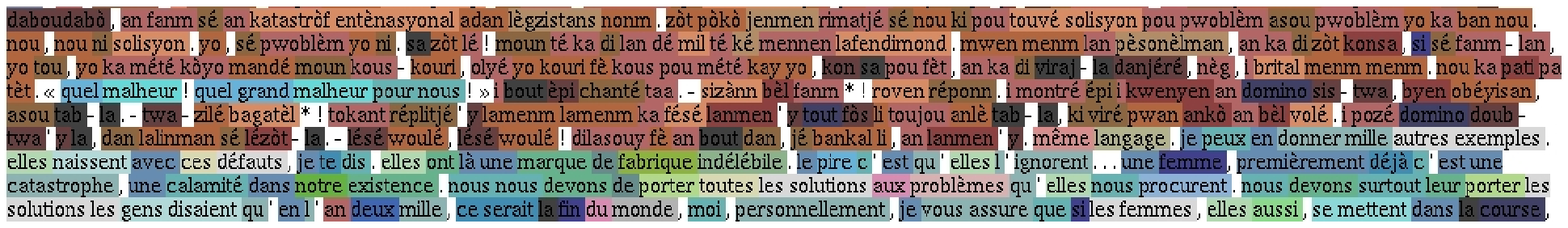,width=7.50cm} & \epsfig{file=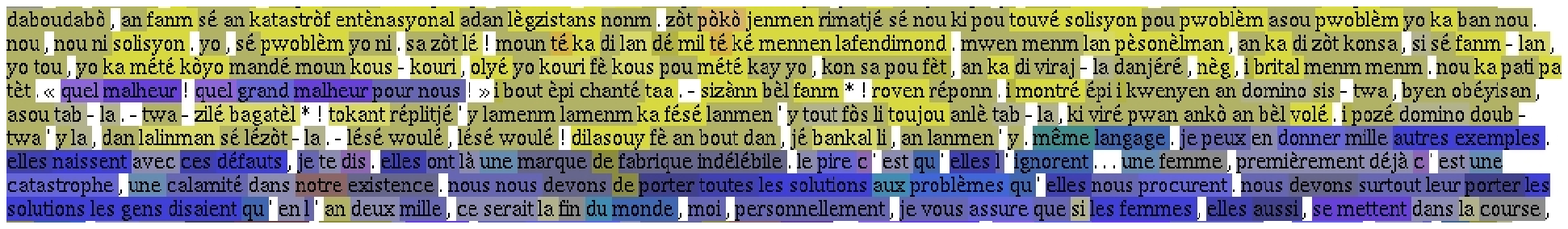,width=7.50cm} \\ \hline
$\pr([{\mathcal{V}}_k]_t | [v_i]_t)$ & $y_{i,k}$\\ \hline
\end{tabular}
\end{center}
\caption{Displaying word coordinates on an RGB space for an extract of
a bilingual novel where French and French-based Creole fragments are
intertwined.\label{f-lavw}}
\end{figure*}

An even more interesting experiment has consisted in trying the
program on texts where languages are more intricately mixed. This is
quite typically so in literature from multilingual regions, like in
the case of the Creole-speaking communities mentioned in the
introduction.  The linguistic situation actually is reflected in the
literature generated in those regions; as an example, we display
(fig. \ref{f-lavw}) an extract of a bilingual novel from a Caribbean
author, where segments in French and Creole alternate. 
In this case, rather than plotting $\pr([v_i]_t | [{\mathcal{V}}_k]_t)$
for soft spectral clustering, we have plotted $\pr([{\mathcal{V}}_k]_t | [v_i]_t)
= \pr([v_i]_t | [{\mathcal{V}}_k]_t) \pr([{\mathcal{V}}_k]_t) / \pr([v_i]_t)$,
using $\pi_i = \pr([v_i]_t)$, and solving the linear system $\bm{p}^\star = 
\tilde{Y}^{-1} \pi$ to find $p^\star_k = \pr([{\mathcal{V}}_k]_t)$ ($k=1, 2, ..., v$).
Since we plot color levels for each word, it should be more convenient
to yield sharper visual differences between languages. While both languages share many words, 
the results display quite surprising contrasts, and these are actually sharper
when plotting $\pr([{\mathcal{V}}_k]_t | [v_i]_t)$.
In the crop of fig. \ref{f-lavw}, the program has
even managed to extract a short French sentence (\textit{quel malheur, quel grand
malheur pour nous}) out of a Creole segment.
\begin{figure}[t]
\begin{center}
\begin{tabular}{c}
\epsfig{file=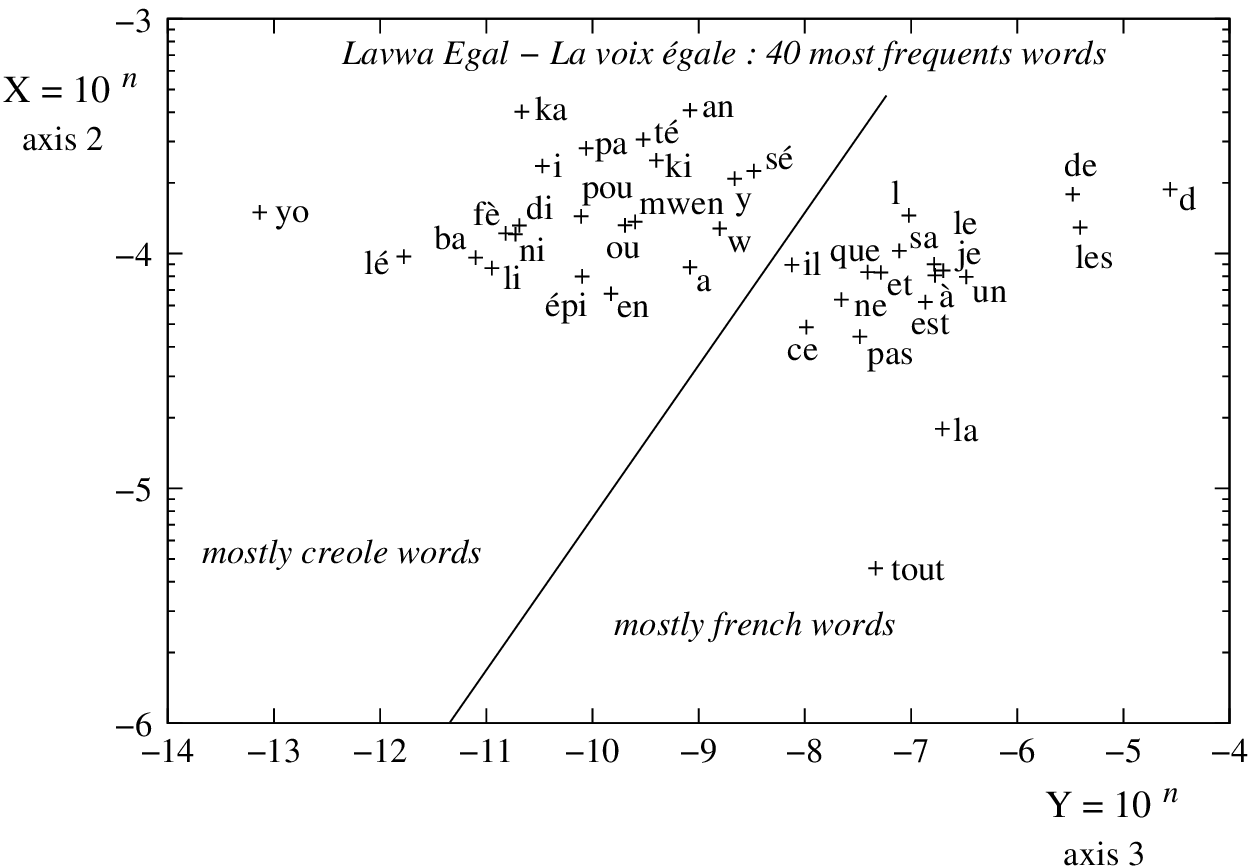,width=7cm}
\end{tabular}
\end{center}
\caption{The 40 most frequent words in a French-Creole Caribbean
novel, projected on a plane along the second and third principal axes. 
\label{f-40mfw}}
\end{figure}
Finally, Figure \ref{f-40mfw} presents a 2D plot on clusters $k=2,3$
of the forty most frequent words. It was interesting to notice that 
two soft clusters were enough to make appear a clear 
frontier between the two languages, though each side of this
frontier obviously contains words that are found on both languages
(\textit{a}, \textit{an}, \textit{la}, \textit{ni}, \textit{ou}, \textit{tout}, \textit{y}, etc.). 

\section{Conclusion}

In this paper, we have provided a new way to build a Markov chain out
of a text, which satisfies all conditions for a convenient spectral
decomposition. We have provided a novel way to interpret the results
of spectral decomposition, in terms of soft clustering. This
probabilistic interpretation allows to avoid the complexity gap that
follows from traditional hard spectral clustering. This brings a
natural approach to process a Markov chain, and make a soft clustering
out of its state space. We bring an extensive comparison of hard and
soft spectral clustering, along with some extended results on
conventional spectral clustering.

The experiments clearly display the potential of such a method. It has
the ability to separate two implicit Markov processes which have
contributed in a mixed proportion to the generation of one single
observable output. The results presented here are obtained from a
simple bigram model; they can be improved, at the cost of some
computation time, by taking into account variable length n-grams. The
property of separating two implicit generation processes has an
obvious application in language identification, and in language
hierarchical clustering, as we have shown.  Moreover, we believe that
it also has a potential to prove useful, with some more research, in
other types of applications like: identifying mixed discourse {\em
genres} (e.g.  formal vs. informal); identifying segments of text with
different topics; or spotting sources within texts of mixed
authorship.

In future works, we plan to drill down our soft spectral clustering
results, to converge towards a complete probabilistic interpretation
of spectral decomposition. Another target of our future research is
using this method on other matrices computed from a text, like the
matrix of distributional similarity (measuring ``paradigmatic''
syntactic distance instead of ``syntagmatic'' syntactic distance, cf.
Introduction), with the aim of clustering syntactic and semantic
categories in loosely-described languages.

\bibliographystyle{ecai2006}
%\bibliographystyle{IEEEbib}
%\vspace{-0.15cm}

\bibliography{bibgen}

\begin{thebibliography}{10}

\bibitem{bjLS}
F.~R. Bach and M.~I. Jordan, `Learning spectral clustering', in {\em NIPS* 16},
  eds., S.~Thrun, L.~Saul, and B.~Schoelkopf. MIT Press, (2003).

\bibitem{bgUV}
M.~Belkhin and J.~Goldsmith, `Using eigenvectors of the bigram graph to infer
  morpheme identity', in {\em Proc.\ of the {ACL'02 Workshop on Morphological
  and Phonological Learning}}, (2002).

\bibitem{bnLE}
M.~Belkhin and P.~Niyogi, `Laplacian eigenmaps and spectral techniques for
  embedding and clustering', in {\em NIPS* 14}, eds., T.~G. Dietterich,
  S.~Becker, and Z.~Ghahramani. MIT Press, (2001).

\bibitem{brcTS}
R.~Besan{\c{c}}on, M.~Rajman, and J.-C. Chappelier, `Textual similarities based
  on a distributional approach', in {\em 10th International Workshop on
  Database And Expert Systems Applications (DEXA'99)}, pp. 180--184. IEEE
  Press, (1999).

\bibitem{cSG}
F.~R.~K. Chung, {\em Spectral {G}raph {T}heory}, volume~92, Regional Conference
  Series in Mathematics, American Mathematical Society, 1997.

\bibitem{dlpSM}
I.~Dagan, L.~Lee, and F.~Pereira, `Similarity-based models of cooccurrence
  probabilities', {\em Machine Learning}, {\bf 34}(1-3),  43--69, (1999).

\bibitem{sCLG}
F.~de~Saussure, {\em Cours de {L}inguistique {G}{\'{e}}n{\'{e}}rale}, Payot,
  Paris, 1916.
\newblock {e}dited by C. Bally and A. Sechehaye. English translation by R.
  Harris: {\em Course in General Linguistics}, London, Duckworth, 1983.

\bibitem{dlrML}
A.~P. Dempster, N.~M. Laird, and D.~B. Rubin, `Maximum likelihood from
  incomplete data via the {EM} algorithm', {\em J. of the Royal Stat. Soc. B},
  {\bf 39},  1--38, (1977).

\bibitem{dAT}
C.~Ding, `A tutorial on {S}pectral {C}lustering', in {\em Tutorials of the
  21$^{~th}$ ICML}, (2004).

\bibitem{gLIS}
G.~Greffenstette, `Comparing two language identification schemes', in {\em
  Journ{\'{e}}s Internationales d'Analyse des Donn{\'{e}}es Textuelles (JADT)},
  pp. 263--268, (1995).

\bibitem{jTC}
T.~Joachims, `Text classification with suport vector machines: Learning with
  many relevant features', in {\em Proc.\ of the 13$^{~th}$ European Conference
  on Machine Learning}, pp. 137--142, (1998).

\bibitem{jTI}
T.~Joachims, `Transductive inference for text classification using support
  vector machines', in {\em Proc.\ of the 16$^{~th}$ International Conference
  on Machine Learning}, pp. 200--209, (1999).

\bibitem{msLS}
M.~Meil{\u{a}} and J.~Shi, `Learning segmentation by random walks', in {\em
  NIPS* 14}, eds., T.~G. Dietterich, S.~Becker, and Z.~Ghahramani. MIT Press,
  (2001).

\bibitem{mSA}
B.~Mohar, `Some applications of {L}aplace eigenvalues of graphs', in {\em Graph
  symmetry: algebraic methods and applications}, eds., G.~Hann and
  G.~Sabidussi,  225--275, NATO ASI Series, (1997).

\bibitem{njwOS}
A.~Y. Ng, M.~I. Jordan, and Y.~Weiss, `On spectral clustering: Analysis and an
  algorithm', in {\em NIPS* 14}, eds., T.~G. Dietterich, S.~Becker, and
  Z.~Ghahramani, Cambridge, MA, (2001). MIT Press.

\bibitem{psCN}
F.~Peng and D.~Schuurmans, `Combining naive bayes and $n$-gram language models
  for text classification', in {\em Proc.\ of the 25$^{~th}$ ECIR}, pp.
  335--350, (2003).

\bibitem{ptlDC}
F.~Pereira, N.~Tishby, and L.~Lee, `Distributional clustering of {E}nglish
  words', in {\em Proc.\ of the 31$^{~st}$ Meeting of the Association for
  Computational Linguistics}, pp. 183--190, (1993).

\bibitem{ssBT}
R.~E. Schapire and Y.~Singer, `{BoosTexter}: A boosting-based system for text
  categorization', {\em Machine Learning}, {\bf 39},  135--168, (2000).

\bibitem{sPOS}
H.~Sch{\"u}tze, `Part-of-speech induction from scratch', in {\em Proc.\ of the
  31$^{~st}$ Meeting of the Association for Computational Linguistics}, pp.
  251--258, (1993).

\bibitem{smNC}
J.~Shi and J.~Malik, `Normalized cuts and image segmentation', {\em IEEE
  TPAMI}, {\bf 22},  888--905, (2000).

\bibitem{uiTS}
M.~Utiyama and H.~Isahara, `A statistical model for domain-independent text
  segmentation', in {\em Proc.\ of the 39$^{~th}$ Meeting of the Association
  for Computational Linguistics}, pp. 491--498, (2001).

\bibitem{vIST}
Trung-Hung Vo, `Construction d'un outil pour identifier et segmenter
  automatiquement un texte h{\'{e}}t{\'{e}}rog{\`{e}}ne en zones
  homog{\`{e}}nes', in {\em Recherche Innovation Vietnam et Francophonie}, pp.
  175--178, (2004).

\end{thebibliography}

\end{document}